\documentclass[runningheads]{llncs}
\usepackage{array}
\usepackage{accsupp}
\PassOptionsToPackage{hyphens}{url}
\usepackage{hyperref}
\usepackage{amsmath}
\usepackage{subfig}
\usepackage{amsfonts}
\usepackage{amssymb}
\usepackage{todonotes}
\usepackage{mathtools}
\usepackage[linesnumbered,ruled,vlined]{algorithm2e}
\usepackage{tikz}
\usetikzlibrary[topaths]
\usetikzlibrary{calc}
\usepackage{balance}
\usepackage{graphicx}
\usepackage{xspace}
\newcount\mycount
\newcolumntype{L}{>{\RaggedRight\arraybackslash}X}
\newcolumntype{Y}{ >{\Centering}X}
\newcommand{\A}{\mathcal{A}}
\newcommand{\C}{\mathcal{C}}
\newcommand{\AS}{\mathcal{AS}}
\newcommand{\W}{\mathcal{W}}

\begin{document}
\title{Trust-based Multiagent Consensus For Weightings Aggregation}
\titlerunning{Trust-based Multiagent Consensus For Weightings Aggregation}
%
\author{Bruno Yun\inst{1} \and
Madalina Croitoru\inst{2}}

%
\authorrunning{Yun and Croitoru}
%
\institute{$^1$ University of Aberdeen, Scotland\\ $^2$ University of Montpellier, France}
\maketitle              
\begin{abstract}
We introduce a framework for reaching a consensus amongst several agents communicating via a trust network on conflicting information about their environment. We formalise our approach and provide an empirical and theoretical analysis of its properties.

\end{abstract}

\section{Introduction}

We place ourselves in the context of a multi-agent setting where the agents need to reach consensus regarding the description of their jointly observed environment.
The applications for this of setting are numerous stemming from autonomous drones to driver-less cars.
Existing work address this problem either using trust mechanisms \cite{ramchurn_trust_2004,gopal_selfish_2016} or using opinion spreading techniques \cite{degroot_reaching_1974}.
However, these approaches make the assumption that the information exchanged by the agents and used to describe the environment is seen as a black box.
This is an unrealistic simplification since, in many of the aforementioned applications, the information exchanged between agents may be inconsistent due to the incomplete nature of the agents' viewpoint.
As a consequence, certain information might weaken others, or, contrary, might reinforce them. Furthermore, the agents communicate amongst each other with various degrees of communication trust. 

We propose to combine two techniques to address the aforementioned problem and study its theoretical and empirical properties. To model the information interaction, we propose to use argumentation-based ranking semantics \cite{besnard_logic-based_2001,leite_social_2011,amgoud_ranking-based_2013,cayrol_graduality_2005,matt_game-theoretic_2008}. These semantics allow for ``attacked'' pieces of information to have a lower ranking than the non-attacked ones. Such ranking semantics are used directly w.r.t.\ their application desirable properties \cite{bonzon_comparative_2016,leite_social_2011,matt_game-theoretic_2008,cayrol_graduality_2005,amgoud_ranking-based_2013}. Then, to model the information spreading based on the trust model, we use Degroot's model \cite{degroot_reaching_1974}. 

Our contributions lie in the proposal and analysis of the first trust-based consensus finding framework based on argumentative ranking semantics.
This advances the state-of-the-art by a finer grained consideration of the exchanged information by the agents. 
The framework is shown to converge under certain conditions. We unveil a property that allows for the modification of the agent trust matrix without ``losing'' the convergence. In our empirical study, we show that the framework is scalable for real world applications.

\section{Aggregation Framework}

\subsection{Background notions}

An argumentation framework (AF) \cite{dung_acceptability_1995} is $\AS = (\A,\C)$ where $\A$ is a set of arguments and $\C \subseteq \A \times \A$ is a set of directed attacks. The set of attackers of an argument $a$ is denoted by $Att(a)$. Namely, $Att(a) = \{ b \in \A$ s.t.\ $(b,a) \in \C\}$. 
A weighting function on $\A$ is a function $w$ that associates a positive number to each element of $\A$, i.e.\ $w: \A \to \mathbb{R}^+$. The set of all possible weighting functions over $\A$ is denoted by $\W_{\A}$. 
A property on a set of weighting functions $\W$ is a function $\rho: \W \to \{0, 1\}$ s.t.\ for every $w \in \W, \rho(w) = 0$ if the property is not satisfied and $\rho(w) = 1$, otherwise. 
The set of all possible properties on $\W$ is defined by $\Omega(\W)$.
These functions (such as void precedence or cardinality precedence) are defined and thoroughly analysed in the literature \cite{bonzon_comparative_2016,leite_social_2011,matt_game-theoretic_2008,cayrol_graduality_2005,amgoud_ranking-based_2013}.
%

\begin{example}
\label{ex:graph-args}
Let $\AS = (\A, \C)$ be an AF representing the arguments exchanged by four traffic drones, with $\A = \{ a,b,c,d,e\}$ and $\C = \{ (a,e), (d,a), (b,a),$ $(e,d),$ $(b,c), (c,e) \}$. We have $Att(a) = \{b,d\}$ and $Att(b) = \emptyset$.
%
%
%
%
Let us consider the following weighting functions $\{w_1, w_2, w_3, w_4,w_5  \} \subset \W_{\A}$ s.t.\ 
$w_1(a) = 0.38, w_1(b) = 1, w_1(c) =  0.5, w_1(d) = 0.65, w_1(e) = 0.53$,
$w_2(a) = 0.07, w_2(b) = 0.91  , w_2(c) = 0.08 , w_2(d) =  0.2, w_2(e) = 0.78$,
$ w_3(a) = 0.2, w_3(b) = 1, w_3(c) = 0.62, w_3(d) =  0.7 , w_3(e) =  0.40$,
$ w_4(a) = 0.17, w_4(b)=  1, w_4(c) =  0.25, w_4(d) = 0.25 , w_4(e) = 0.5$,
 $w_5(a) =  1, w_5(b) =  0.21, w_5(c) = 0.75 , w_5(d) =  0.32$ and $w_5(e) = 0.40$. 

$\rho_{void}$ is satisfied by $w_1, w_2, w_3$ and $w_4$ but not by $w_5$.
$\rho_{card}$ is only satisfied by $w_3$ and $\rho_{self}$ is satisfied by $w_1, w_2, w_3, w_4$ and $w_5$.

\end{example}


\subsection{Framework's Elements}

We now introduce a three step framework for $k$ agents in $Ind = \{ A_1, A_2, \dots, A_k\}$ to reach a consensus about what can be the possible weights of the arguments in an AF $\AS = (\A,\C)$.
 
\begin{enumerate}
\item \textbf{(Deciding the Weighting Library)}
Once the agents agree on a set of desirable properties of the weighting functions $\Omega^* \subseteq \Omega(\W_{\A})$, we use the set of properties $\Omega^*$ for obtaining the filtered set $\W_1 \subseteq \W_{\A}$ that satisfies all of the properties in $\Omega^*$.  The set $W_1$ is called the \textit{weighting library} w.r.t.\ $\W_{\A}$ and $\Omega^*$. The weighting library is populated with weighting functions obtained from ranking-based semantics satisfying the selected properties.

\item \textbf{(Score assignment)}
The agents have to make two choices. First, via the function $P: Ind \to 2^{\W_1}$ that takes as input an individual $A_i \in Ind$ and returns the set of weighting functions considered by $A_i$. Each agent considers a subset of the weighting library $\W_1$. The set of weightings considered by the agents $Ind$ is denoted by $M = \bigcup_{A_i \in Ind} P(A_i)$.
Second, each agent $A_i$ associate a score $0 < s_w \leq 1$ to each of the weighting function $w$ considered (i.e.\ to each element of $P(A_i)$) s.t.\ $\sum_{w \in P(A_i)} s_w =1$. We can thus associate a scoring function $S_i: M \to [0,1]$, to each agent $A_i$ over the weighting function of $M$, defined as for every $A_i \in Ind$ and every $w \in M$: $S_i(w) = s_w$ if $w \in P(A_i)$ and $S_i(w) =0$ otherwise.


\begin{example}[Cont'd Example \ref{ex:graph-args}]
\label{ex:weigting-library}
Let $Ind = \{ A_1, A_2, A_3, A_4 \}$ be a set of four agents. Suppose that the agents of $Ind$ agreed on $\Omega^* = \{\rho_{self}, \rho_{void} \}$, then $\W_1$ w.r.t.\ $\W_{\A}$ and $\Omega^*$ contains $w_1,$ $w_2,$ $w_3,$ $w_4$ but not $w_5$ since $\rho_{void}(w_5) = 0$. 

Assume now that $P(A_1) = \{ w_1, w_2, w_3 \}$, $P(A_2) = \{ w_2, w_3 \}$, $P(A_3) = \{ w_3, w_4 \}$ and $P(A_4) = \{ w_2, w_3, w_4\}$. The set of weightings considered by the agents is  $M=\{ w_1, w_2, w_3, w_4 \}$. In the rest of this paper, we consider $S_1,$ $S_2,$ $S_3$ and $S_4$ depicted in Table \ref{tabl:f4}.

\begin{table}
\centering
\parbox{.45\linewidth}{
\centering
\begin{tabular}{|c|c|c|c|c|}
\hline
& $w_1$ & $w_2$ & $w_3$ & $w_4$ \\
\hline
$S_1$ & 0.4 & 0.3 & 0.3 & 0\\
$S_2$ & 0 & 0.5 & 0.5 & 0\\
$S_3$ & 0 & 0 & 0.7 & 0.3\\
$S_4$ & 0 & 0.2 & 0.1 & 0.7\\
\hline
\end{tabular}
\caption{The scoring function $S_1, S_2, S_3$ and $S_4$.}
\label{tabl:f4}
}
\quad
\parbox{.45\linewidth}{
\centering
\begin{tabular}{|c|c|c|c|c|}
\hline
$Ind$ & $A_1$&  $A_2$ & $A_3$ & $A_4$ \\
\hline
$A_1 $ & 0.75 & 0.15 & 0.1 & 0 \\
\hline
$A_2$ & 0.2 & 0.7& 0.1 & 0 \\
\hline
$A_3$ & 0.35 & 0.15 & 0.5 & 0 \\
\hline
$A_4$ & 0.3 & 0.3 & 0.3& 0.1 \\ 
\hline
\end{tabular}
\caption{Agent trust matrix $V$}
\label{tabl:influences}}
\end{table}

\end{example}

\item \textbf{(Preference Propagation)} This phase takes into account the trust between agents to reach a consensus on the weightings. We define the trust of agent $A_j$ for agent $A_i$ by the number $v_{i,j}$ with $\sum_{j=1}^{j=k} v_{i,j} = 1$ (with $v_{i,j}=0$ either when agent $A_i$ does not know of the existence of agent $A_j$ or when $A_i$ does not trust the opinion of $A_j$). One way of representing the trust between the agents is a $k \times k$ agent trust matrix $V$ (see Table \ref{tabl:influences}).


\end{enumerate}

\subsection{Weightings Aggregation}

Within this framework, we can chose many propagation methods. In this paper, due to its large popularity and simplicity, we follow the DeGroot model \cite{degroot_reaching_1974} where the scoring functions are updated over time so that $S_w^{(n)} = V \times S_w^{(n-1)} = V^n \times S_w^{(0)}$, where $S_w^{(0)}$ is the column vector $(S_1(w), \dots, S_k(w))^T$.
A consensus is reached iff for every $w \in M,$ all the $k$ elements of $S^{(n)}_w$ converge toward the same limit as $n \to +\infty$ \cite{golub_naive_2010,berger_necessary_1981} and is computed by finding a vector $\pi = (\pi_1, \dots, \pi_k)$ s.t.\ $\sum_{i=1}^{k} \pi_i= 1, \pi V = \pi$ and for every $i \in \{1, \dots, k \}, \pi_i \geq 0$. The score function $S^*$ of all the agents in the consensus is $S^*(w) = \sum_{i=1}^{i=k} S_i(w) \times \pi_i$.

\begin{example}[Cont'd Example \ref{ex:weigting-library}]
\label{ex:convergence}
The initial score for $w_1$ across all the agents is the vector $S^{(0)}_{w_1} = (S_1(w_1),$ $S_2(w_1),$ $S_3(w_1),$ $S_4(w_1))^T = (0.4,0,0,0)^T$. In the agent trust matrix, $A_1$ weights himself more than agents $A_2$ and $A_3$ and ignores agent $A_4$. We get that the updated scores for $w_1$ is $S_{w_1}^{(1)}= V \times S_{w_1}^{(0)} = (0.3, 0.08, 0.14, 0.12)^T$.
%
%
Multiple iterations of this process leads to a consensus on the score of $w_1$, i.e.\
$S_{w_1}^{(n)} = V^n \times S_{w_1}^{(0)} \rightarrow (0.2, 0.2, 0.2, 0.2 )^T$.
In our example, there is a unique vector $\pi = (\frac{1}{2}, \frac{1}{3}, \frac{1}{6}, 0)$ s.t.\ $\pi V = \pi$. We have $S^* (w_1) = \frac{1}{2} \times 0.4 = 0.2$, $S^*(w_2) = \frac{1}{2} \times 0.3 + \frac{1}{3} \times 0.5 \approx 0.32$, $S^*(w_3) = \frac{1}{2} \times 0.3 + \frac{1}{3} \times 0.5 + \frac{1}{6} \times 0.7 \approx 0.43$ and $S^*(w_4) = \frac{1}{6} \times 0.3 = 0.05$.

\end{example}

\subsection{Practical Use Case}
Before proceeding to analyse the framework, we provide a practical use case within the traffic monitoring inspired example. 
We consider three driver-less cars (A, B and C) that are simultaneously arriving at a very narrow roundabout. The roundabout is monitored by four traffic drones. Each of the traffic drones has different opinions on the weights of the arguments depending on their geographical positions. For instance $A_1$ and $A_2$ are drones that both have high-ground positions and a clear view of the roundabout, $A_3$ is a drone with a ground level vision and $A_4$ has an obstructed vision of the situation. $A_1$ and $A_2$ are confident about their own judgement whereas $A_4$ has low trust about his own opinion. Moreover, $A_1, A_2, A_3$ can communicate with each others whereas the opinion of $A_4$ is not taken into account by the other agents.
The drones, depending on their position, attempt to model the following conflicting arguments:
$\textbf{(a)}$ A will enter the roundabout first and it means that B will not enter.
$\textbf{(b)}$ There is a pedestrian crossing in front of the car A and the speed of car B is fast which means that car A will not enter the roundabout first and that the speed of B is not slow.
$\textbf{(c)}$ The speed of car B is slow which means that B will not enter first.
$\textbf{(d)}$ C will enter the roundabout first and it means that A will not enter.
$\textbf{(e)}$ B will enter the roundabout first and it means that C will not enter.
 
Our framework will output a consensus on the weights allowing to have an unified view of the traffic based on their agent trust network.
Concretely, if we consider the agent trust matrix in Table \ref{tabl:influences} and the initial opinions of the drones in Table \ref{tabl:f4} about the weighting functions of Example \ref{ex:graph-args}, our framework will output an optimal subset of the weightings (in this case $\{w_3\}$) or an aggregated weighting function $w^*$ (see Definition \ref{def:aggregated}).


\section{Framework Analysis}

In this section, we provide a theoretical and empirical evaluation of our framework. The theoretical analysis of the proposed framework and our results will allow to answer the following research questions: 
%
%
(1) Can agent modify their trust network without losing the convergence?
(2) What are the semantics properties of the weights outputted by the consensus?
(3) Is this approach scalable for a large number of agents and/or arguments?

\subsection{Theoretical Properties}

We first address the question: Can agents modify their trust network without losing the convergence? 
The previous work by \cite{golub_naive_2010,berger_necessary_1981} proposed many conditions for a trust network to achieve a consensus.
In the next proposition, we show how the agent trust matrix can be revised without losing the convergence. 

\begin{proposition}
\label{prop:revised-consensus}
Let $V$ be a trust matrix s.t.\ a consensus can be reached, $V'$ be right stochastic matrix and $F: \mathbb{R}^{k\times k} \to \mathbb{R}^{k\times k}$ be a matrix function s.t.\:

\begin{multline*}
V = \begin{pmatrix} 
v_{1,1} & \dots & v_{1,n} \\
\vdots & \ddots & \vdots \\
v_{k,1} & \dots & v_{k,k}
\end{pmatrix}= 
\begin{pmatrix} 
f_{1,1}(v'_{1,1}) & \dots & f_{1,n}(v'_{1,n}) \\
\vdots & \ddots & \vdots \\
f_{k,1}(v'_{k,1}) & \dots & f_{k,k}(v'_{k,k})
\end{pmatrix}= F(V')
\end{multline*}

where $f_{i,j}: [0,1] \to [0,1]$ is s.t.\ for every $v \in [0,1], f_{i,j}(v) = 0$ if $v=0$ and $f_{i,j}(v) = v_{i,j}$, otherwise. It holds that a consensus can be reached with $V'$.

\end{proposition}

We now show how the scoring function $S^*$ obtained from the consensus between agents can be used for choosing/creating a weighting function for an AF.
%
%
We first define the output of the consensus as an optimal set of weighting functions that should be used for an AF.

\begin{definition}[Output of a consensus]
The output of the consensus is the set of weightings $M^* \subseteq M$ s.t.\ for every $w, w' \in M^*$ and $w'' \in M \setminus M^*$, $S^*(w) = S^*(w')$ and $S^*(w') > S^*(w'')$.
\end{definition}

\begin{example}[Cont'd Example \ref{ex:convergence}]
The output of the consensus is the set of weightings $M^*=\{w_3\}$. $w_3$ is thus the weighting that will be used for the traffic AF.
\end{example}

The shape of the agent trust network determines the influence of each agents and the agents that are not communicating with the only strongly connected component have no influence in the consensus.

\begin{proposition}
If the agent trust network is aperiodic and there is only one strongly connected component $S\subseteq Ind$ that is closed (no edges to the agents outside of the strongly connected component) then for every $A_i \in S, \pi_i >0$ and for every $A_j \in Ind \setminus S, \pi_j = 0$.
\end{proposition}

Furthermore, we can see that an agent that possesses more than half of the total influence of the network can force his opinion to the other agents.

\begin{proposition}
If there exists an agent $A_i \in Ind$ s.t.\ $\pi_i > 0.5$ then it is possible for $A_i$ to choose $w \in M$ s.t.\ $w \in M^*$.
\end{proposition}


The advantage of using this approach is that the output of the consensus is a subset of $M$ and will thus satisfy every property of $\Omega^*$. However, the cons are that the output of the consensus is not always a single weighting function and, in the case where an agent has most of the influence, he can force the choice of one weighting function.
Note that \cite{golub_naive_2010} introduced a condition on the agent trust matrix so that the consensus is ``wise'', i.e.\ there is no agent that retain too much influence in the consensus. 

We now suppose that instead of using the scoring function $S^*$ for choosing the best weighting functions in the consensus, we use the scoring function $S^*$ for obtaining an aggregated weighting function.

\begin{definition}[Aggregated weighting function]
The aggregated weighting function w.r.t.\ $S^*$ is $w^*: \A \to \mathbb{R}$ s.t.\ for every $a \in \A, w^*(a) = \sum\limits_{w \in M} S^*(w) \times w(a)$. 
\label{def:aggregated}
\end{definition}

\begin{example}[Cont'd Example \ref{ex:converge-limit-scoring}]
The aggregated weighting function is $w^*$ s.t.\ $w^*(a) = 0.38 \times 0.2 + 0.07 \times 0.32 + 0.2 \times 0.43 + 0.17 \times 0.05 \approx 0.19$, $w^*(b) = 0.97, w^*(c) = 0.40, w^*(d) = 0.51$ and $w^*(e) = 0.55$.



\end{example}

\begin{proposition}
Let $a,b \in \A$ be two arbitrary arguments of $\A$. If for every weighting function $w \in M$, it holds that $w(a) > w(b)$ then $w^*(a) > w^*(b)$. 
\label{prop-aggreg}
\end{proposition}


If all the weighting functions chosen by the agents satisfy some properties then the aggregated weighting function also satisfies the same properties.

\begin{proposition}
Let us consider the weighting function $w \in M$. It holds that (1) $\rho_{void}(w) = 1$ then $\rho_{void}(w^*) =1 $, (2) $\rho_{card}(w) = 1$ then $\rho_{card}(w^*) =1 $ and (3) $\rho_{self}(w) = 1$ then $\rho_{self}(w^*) =1 $
\label{prop:generalisation}
\end{proposition}

Proposition \ref{prop:generalisation} is important as it shows that the aggregated weighting function $w^*$ keeps the desirable properties of the weighting functions. For readability purposes, we did not include all of the properties but this proposition can be generalised for most of the ranking-based semantics properties of \cite{bonzon_comparative_2016} (quality precedence, counter-transitivity, strict counter-transitivity, defense precedence, distributed defense precedence, etc.).
Please note that it is not true that $w^*$ will satisfy every property of $\Omega^*$ (see Example \ref{ex:counter-ex-omega} for a counter-example).

\begin{example}
\label{ex:counter-ex-omega}

Suppose that $\AS$ has three arguments $a,b$ and $c$, $M = \{ w'_1, w'_2 \},$ $S^*(w'_1) = 0.5, S^*(w'_2) = 0.5$ and $w'_1, w'_2$ are s.t.\ $w'_1(a) = 1, w'_1(b) = 0.5, w'_1(c) = 0$, $w'_2(a) = 0$, $w'_2(b) = 0.5$ and $w'_2(c) = 1$.
If we consider the property \emph{All Different} defined as $\rho_{diff}(w)=1$ iff for every $a,b \in \A, w(a) \neq w(b)$. It holds that $\rho_{diff}(w'_1) = \rho_{diff}(w'_2) = 1$ but $\rho_{diff}(w'^*) = 0$.
\end{example}

The pros of using this approach are that the ranking of arguments is ``preserved'' which leads to the satisfaction of most argumentation properties by the aggregated weighting function $w^*$ and that the aggregated weighting $w^*$ function is unique. However, the disadvantage is that $w^*$ can violate the properties of $\Omega^*$.

\subsection{Empirical Evaluation}

Let us analyse the time and the number of steps needed in practice to achieved the consensus (with a precision $\varepsilon$). 
We implemented an algorithm that takes as input the agent trust matrix $V$ and a precision $\varepsilon$ and computes $V' = V^K, K \geq 1,$ s.t.\ all the lines of $V'$ are nearly-equal with a precision $\varepsilon$, i.e. for every two lines $V'_1$ and $V'_2$ of $V'$, ${V'_1}_i$ the $i$-th element of $V'_1$, ${V'_2}_i$ the $i$-th element of $V'_2$, we have $|{V'_1}_i - {V'_2}_i| \leq \varepsilon$.
Then, since all the lines of $V'$ are nearly-equal with a precision $\varepsilon$, we retrieve the first line of $V'$ which is an approximation of the vector $\pi$. The algorithm returns the vector corresponding to $S^*$. 

%
%
%
%

\subsubsection{Data gathering}
\label{sec:generation-mat}
The three-step procedure to follow to generate a random $k \times k$ matrix $V$ that can reach a consensus is: (1) Pick $k$ numbers $n_1, \dots, n_k$ between $0$ and $k$ that we divide by $\sum_{i=1}^{k} n_i$. We thus obtain $n'_1, \dots, n'_k$ where $n'_i = n_i \times \frac{1}{\sum_{i=1}^{k} n_i}$. Note that we make sure that not all the $n_i$ are equal to zero. (2) The numbers $n'_1, \dots, n'_k$ are assigned to the first line of $V$. The step 1 is repeated to obtain all $k$ lines of $V$. At this point, the matrix $V$ is right stochastic, i.e.\ the sum of the elements of each line is $1$. (3) For a fixed positive integer $c$, we check whether any of the matrices $V, V^2, \dots, V^c$ have at least one column s.t.\ every element in that column is positive. If yes, the matrix $V$ can reach a consensus. If not, we repeat the steps (1) and (2).

\subsubsection{Experimental results}

We generated square matrices of increasing sizes (from 50 to 2000 with a step of 50) and experimented with two different precisions ($10^{-3}$ and $10^{-5}$). We repeated each run five times to obtain the average time and average number of steps needed to get the vector $\pi$. The results are displayed in Figure \ref{fig:steps-agents} and \ref{fig:time-agent}.
All experiments presented in this section were performed on a Mac machine running with one allocated processor (100\%) of an Intel core i5 2.80 GHz and 8GB of RAM. The all-in-one program for generating and experimenting as well as the results are available online at \url{https://www.dropbox.com/sh/jp62ve810dy3dow/AABKPS3XE-En73_CR4ZeLXsKa?dl=0}.

We can make the following observations: (1) The number of steps needed to reach a consensus decreases with the increase of agents, (2) decreasing the precision (from $10^{-5}$ to $10^{-3}$) decreases the number of steps needed to reach a consensus, (3) decreasing the precision (from $10^{-5}$ to $10^{-3}$) also decreases the total time needed to reach the consensus and (4) the time to reach a consensus follows a polynomial curve w.r.t. the number of agents. This is not surprising as the number of matrix multiplication stays relatively low and the bound for complexity of square matrix multiplication is around $O(n^{2.73})$ \cite{davie_improved_2013,gall_powers_2014}.

\begin{figure}
\subfloat[Average number of steps for reaching a consensus w.r.t.\ the number of agents]{
\begin{tikzpicture}[y=0.8cm, x=0.0020cm,font=\sffamily]
	\draw (0,0) -- coordinate (x axis mid) (2050,0);
    	\draw (0,0) -- coordinate (y axis mid) (0,4);
    	\foreach \i[evaluate= \i as \x using \i+50] in {0,500,...,2000}
     		\draw (\x,1pt) -- (\x,-3pt)
			node[anchor=north] {\x};
			
    	\foreach \y in {0,1,...,4}
     		\draw (1pt,\y) -- (-3pt,\y) 
     			node[anchor=east] {\y};

	\node[below=0.8cm] at (x axis mid) {Number of agents};
	\node[rotate=90, yshift=0.7cm] at (y axis mid) {Average number of steps};

	\draw plot[mark=*, mark options={fill=white}] 
		file {div_soft.data};
	\draw plot[mark=triangle*, mark options={fill=white} ] 
		file {div_ciu.data};
		    
	\begin{scope}[xshift=2cm, yshift=3cm] 
	\draw (0,0) -- 
		plot[mark=*, mark options={fill=white}] (0.25,0) -- (0.5,0) 
		node[right]{$\varepsilon = 10^{-5}$};
	\draw[yshift=\baselineskip] (0,0) -- 
		plot[mark=triangle*, mark options={fill=white}] (0.25,0) -- (0.5,0)
		node[right]{$\varepsilon = 10^{-3}$};
	\end{scope}
	\end{tikzpicture}
		\label{fig:steps-agents}}
\quad
\subfloat[Average time needed to reach a consensus as a function of the number of agents]{
\begin{tikzpicture}[y=0.00035cm, x=0.0020cm,font=\sffamily]
	\draw (0,0) -- coordinate (x axis mid) (2050,0);
    	\draw (0,0) -- coordinate (y axis mid) (0,12000);
    	\foreach \i[evaluate= \i as \x using \i+50] in {0,500,...,2000}
     		\draw (\x,1pt) -- (\x,-3pt)
			node[anchor=north] {\x};
			
    	\foreach \y in {0,1000,...,12000}
     		\draw (1pt,\y) -- (-3pt,\y) 
     			node[anchor=east] {\y};

	\node[below=0.8cm] at (x axis mid) {Number of agents};
	\node[rotate=90, yshift=1.2cm] at (y axis mid) {Time (ms)};

	\draw plot[mark=*, mark options={fill=white}] 
		file {a.data};
	\draw plot[mark=triangle*, mark options={fill=white} ] 
		file {b.data};
		    
	\begin{scope}[xshift=2cm, yshift=3cm] 
	\draw (0,0) -- 
		plot[mark=*, mark options={fill=white}] (0.25,0) -- (0.5,0) 
		node[right]{$\varepsilon = 10^{-5}$};
	\draw[yshift=\baselineskip] (0,0) -- 
		plot[mark=triangle*, mark options={fill=white}] (0.25,0) -- (0.5,0)
		node[right]{$\varepsilon = 10^{-3}$};
	\end{scope}
	\end{tikzpicture}
\label{fig:time-agent}}
\end{figure}

\section{Discussion}

We introduced a framework for reaching a consensus in a multi-agent setting where agents are arguing about the description of their environment using weighting functions. 
The consensus is used to achieve the selection of a subset of the existing weighting functions that are considered or the aggregation of some of the weighting functions.
We provided a mapping to Euclidian space that allows to illustrate the convergence and the distance to the consensus and a property that allows for the modification of the agent trust matrix that conserves the convergence. Since our work is directly applicable for driver-less cars we have also provided an empirical evaluation to test the scalability of the proposed approach. This evaluation is performed on synthetically generated data and future work is currently directed over the practical implementation of such scenario within a multidisciplinary joint project.


\newpage

\bibliographystyle{splncs04}
\bibliography{bruno}

\end{document}